# 3D Visualization and Spatial Data Mining for Analysis of LULC Images

Kodge B. G.

Department of Computer Science, Swami Vivekanand Mahavidyalaya, Udgir, Latur, India

kodgebg@hotmail.com

*Abstract: The present study is an attempt made to create a new tool for the analysis of Land Use Land Cover (LUCL) images in 3D visualization. This study mainly uses spatial data mining techniques on high resolution LUCL satellite imagery. Visualization of feature space allows exploration of patterns in the image data and insight into the classification process and related uncertainty. Visual Data Mining provides added value to image classifications as the user can be involved in the classification process providing increased confidence in and understanding of the results. In this study, we present a prototype of image segmentation, K-Means clustering and 3D visualization tool for visual data mining (VDM) of LUCL satellite imagery into volume visualization. This volume based representation divides feature space into spheres or voxels. The visualization tool is showcased in a classification study of high-resolution LULC imagery of Latur district (Maharashtra state, India) is used as sample data.*

*Keywords: Visual Data Mining, K-Means Clustering, Knowledge Discovery, LULC Satellite Image, GIS, 3D Space Feature Plot.*

## I. Introduction

Image classification based on satellite imagery is a widely used technique for extracting thematic information on land cover. This image processing step is the translation from spectral reflectance or digital numbers (DN) to thematic information. Classification is often performed to generalize a complex image into a relatively simple set of classes. A classified map is then used as input into a geographic information system (GIS) for further processing or analysis. Such inference is most often less than perfect and there is always an element of uncertainty in a classification result. As it can affect further processing steps and even decision making, it is important to understand, quantify and visualize the classification process [1].

Visual Data Mining (VDM) is a powerful tool which is often overlooked in favour of traditional purely non-visual data mining. It is defined as the process of (semi-)automatically discovering meaningful patterns in data. VDM uses visual interaction to allow a human user to visually extract and explore patterns in data. When conducting a non-visual data mining, no matter how unbiased it may seem, the fact is that by simply choosing to carry out an automated analysis a priori assumptions have been made about what form the important results will take before analysis has actually begun. By visually mining the data, this prior bias can be removed. Whilst the bias is removed, subjectivity of the analysis is increased as it is based on a user's perception, a point highlighted by many machine learning purists. However, this increased subjectivity is compensated for by a vastly increased degree of confidence in the analysis [2]. VDM not only seeks to allow a human user to visually mine data but also to augment the non-visual data mining process. This augmentation usually takes the form of making the automated process more transparent to the user, hence providing increased confidence.

VDM is not commonly applied in remote sensing applications. A traditional supervised remote sensing classification starts with a selection of training pixels or areas that represent specific land cover classes. The spectral and statistical properties of these pixels are then used to classify all unlabelled pixels in the image with a classification algorithm such as the widely used maximum likelihood classifier (commonly implemented in commercial remote sensing software). The accuracy of the classified map is tested with reference pixels that are not used in the training stage. Accuracy assessment usually takes the form of an error matrix with derived accuracy values such as the overall accuracy and the Kappa statistic. Although the error matrix provides an overall assessment of classification accuracy, it does not provide an indication of the spectral dissimilarity of class clusters, uncertainty related to the attribution of class labels to individual pixels, or the spatial distribution of classification uncertainty. In this study, we argue that VDM is an important tool for visual exploration of the data to improve insight into the classification algorithm and identify sources of spatial and thematic uncertainty. Recent studies showed that exploratory visualization tools can help to improve the image analyst's understanding of uncertainty in a classified image scene. They proposed a combination of static, dynamic and interactive visualizations for exploration of uncertainty in the classification results. Keim D. A. 2002 developed a visualization tool that allowed for visual interaction with the parameters of a fuzzy classification algorithm [2]. The study showed that visualization of a fuzzy classification algorithm in a 3D feature space plot dynamically linked to a satellite image improves a user's understanding of the sources and locations of uncertainty [1].

## II. Methods, Material And Algorithm

In this study, we developed and present a new VDM prototype to visualize irregular shapes of class clusters and their spectral overlap in a 3D feature space plot. The tool helps to identify the shape of class clusters and the overlap of these class clusters in 3D feature space to highlight sources of uncertainty in the training data for a spectral image classifier. To showcase the visualization prototype we present a classification study based on high-resolution LULC imagery of Latur district (sample image) to assess the value of VDM in semi-automated image classification. The LULC of a particular province may have n number of color combinations for n number of covered regions. The Figure 1 showing the numbered labels of different covered regions. The label 1 in Figure 1 shows the LULC null data (i.e. image background), label 2 (Red) showing the constructed area, label 3 (Green) showing vegetation, label 4 (Blue) showing ground water, label 5 (Yellow) showing the agricultural land, label 6 (Gray) showing the rocky/barren land, and the label 7 (Brown) showing the scrub land respectively [5].

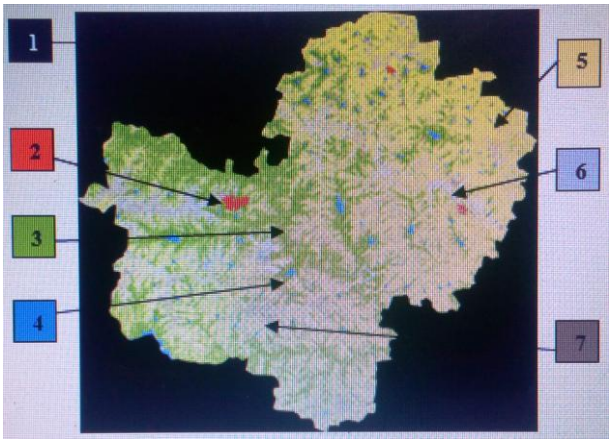

Fig. 1. LULC Image of Latur District (Courtesy NRSC (ISRO), Hyd.)

This study is limited to a pixel-based classification approach; however, the visualization tool can be used for object-oriented classification as well.

The Figure 2 highlights the steps of processes required to classify, segment, cluster, area computation, color (pixel) computation and size computations for 3D visualization of clustered color spheres of LULC image. The proposed algorithm is designed in MATLAB 7.0 and tested successfully on different LULC images for accuracy assessment.

The proposed tool for VDM first read the specified LULC image (RGB) of any province and will ask users to select the number regions (an user interface is provided for selecting colored pixels from LULC image) in which the user would like to segment, classify and visualize. While selecting the number of colored regions (pixels), the algorithm requests users to select the first color (pixel) should be the image background (If the LULC has the image background), it will helps to classify the null data from LULC image. Further the algorithm will process LULC image and starts to segment the processed LULC image.

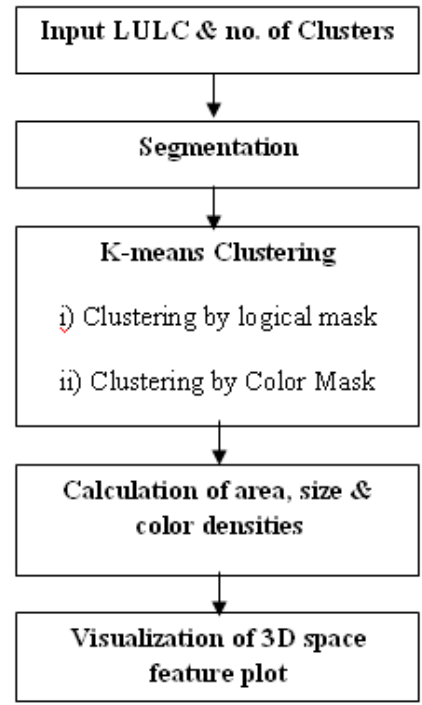

Fig. 2. Algorithm

After segmenting the processed image the tool will generate the segmented image and display on screen with the selected colors as color map at right side. The Figure 3 showing the segmented image of processed LULC image.

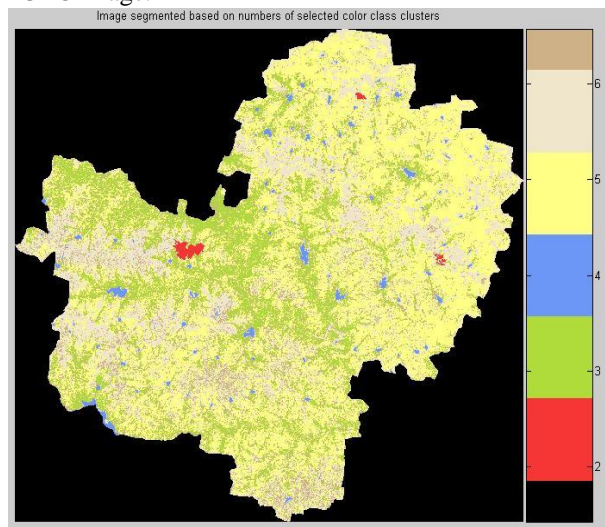

Fig. 3. Segmented LULC Image and Their Color Class Clusters (Color Map/Legend).

Once the segmentation process completed successfully, the toll will starts to cluster the color classes (selected color regions) from image using K-Means clustering technique.

The k-means clustering is a method of cluster analysis which aims to partition n observations into k clusters in which each observation belongs to the cluster with the nearest mean. It is similar to the expectation-maximization algorithm for mixtures of Gaussians in that they both attempt to find the centers of natural clusters in the data as well as in the iterative refinement approach employed by both algorithms. The given a set of color observations (x1, x2, ..., xn), where each observation is a d-dimensional real color class region in LULC, k-means clustering aims to partition the n observations into k sets (k ≤ n) S = {S1, S2, ..., Sk} so as to minimize the within-cluster sum of squares (WCSS):

$$\arg \min_{\mathbf{S}} \sum_{i=1}^{k} \sum_{\mathbf{x}_j \in S_i} \|\mathbf{x}_j - \boldsymbol{\mu}_i\|^2$$

where μi is the mean of points in Si.

After clustering different color class clusters from LUCL, the tool will generates xn number of logical masked individual images matrices and xn number of colored masked individual image matrices and one shared color map.

In this study we have processed 7 color class clusters (Figure 1), so the tool will generates seven logical (black/white) masked and 7 colored masked images matrices and their associated color masks of each selected region of LULC and stores into MATLAB workspace. The generated image matrices and their associated color maps are stored in MATLAB workspace files are shown in Table I.

Table I. Generated Clustered Workspace Files in MATLAB

| Logical Images (B&W) | RGB Color Maps | Clustered Image |
|---|---|---|
| clusterMask1 clusterMask2 clusterMask3 clusterMask4 clusterMask5 clusterMask6 clusterMask7 | IndividualColorMap1 IndividualColorMap2 IndividualColorMap3 IndividualColorMap4 IndividualColorMap5 IndividualColorMap6 IndividualColorMap7 clusteredMap | Clustered |

Here, there are two ways to access the individual color masks (image matrices) of their respective clusters map in this experiment. The first, create 7 number of clustered image matrices which share common color map. The second way is to create 7 numbers of color maps and share common clustered image matrix as shown in Figure 4.

We preferred the second method to create individual color masks of their respective color class clusters, because the sample experiment used an LULC image (RGB) of Latur district having size 563x613x3, and the image matrix for that converted indexed image will be 563x613 and 7x3 respectively (for seven clusters). The numbers of clusters used in the above experiment are seven, so the seven indexed images of size 563x613 are needed to create and it will take 6.99 times (for this experiment) more memory related to the second method as shown in the Table 2.

Table II. Individual Color Masking Methods for an Indexed Image (s)

| Methods | Indexed Image (s) | Color Map (s) | Required Size | Total Size |
|---|---|---|---|---|
| Method 1 | 7 | 1 | 563x613x7=415833+21= and 7x3 | 2415854 |
| Method 2 | 1 | 7 | 563x613 =345119+147= and 7x3x7 | 345266 |

The second method is to create seven color maps instead of seven image matrix. The seven color maps will be having only 7x3 size. In Figure 4, we can see the seven maps (7 cluster example) map1,2,...7 are representing for Individual Color Maps of that respective color clusters as shown in Table 1. And the clustered image matrix ('Clustered' in Table 1) is an indexed image which is associated to their color indices of Individual Color Maps.

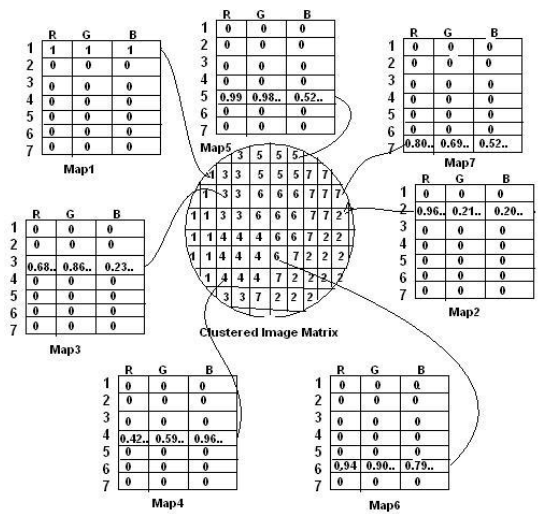

Fig. 4. Individual Color Maps for Clustered Image Matrix

Now one can execute the required single color region mask using the following commands in MATLAB.

>> imshow(clustered, IndividualColorMap1)
>> imshow(clustered, IndividualColorMap2)

>> imshow(clustered, IndividualColorMap3).... upto n number of clusters.

The next process is to compute and store the area of an individual color class cluster. The textual as well as the following bar chart will demonstrate the computed percent area occupied by an individual color class cluster (including cluster1).

Total image area= 345119 pixels

Background area= 156877 pixels or 45.46% image area (i.e Cluster1/null data)

Total LULC area= 188242 pixels or 54.54% image area.

Cluster2 area= 616 pixels or 0.33% image area

Cluster3 area= 32849 pixels or 17.45% image area

Cluster4 area= 2123 pixels or 1.13% image area

Cluster5 area= 88743 pixels or 47.14% image area

Cluster6 area= 39960 pixels or 21.23% image area

Cluster7 area= 23951 pixels or 12.72% image area.

The Figure 3 shows color map for selected individual color class clusters.

## III. RESULTS

The Figure 5 showing the logical image of masked LULC region of Figure 1 (i.e. null data removed).

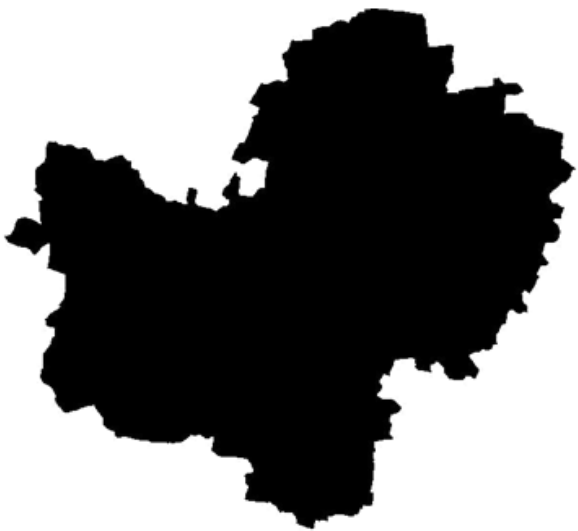

Fig. 5. Logical Image Mask for Cluster 1

After clustering and generated individual color masks images of each selected region(s), the tool will show the computed area wise values (%) through displaying a bar graph of all selected color class clusters. The Figure 6 showing the generated bar graph of covered area wise percent values of clusters.

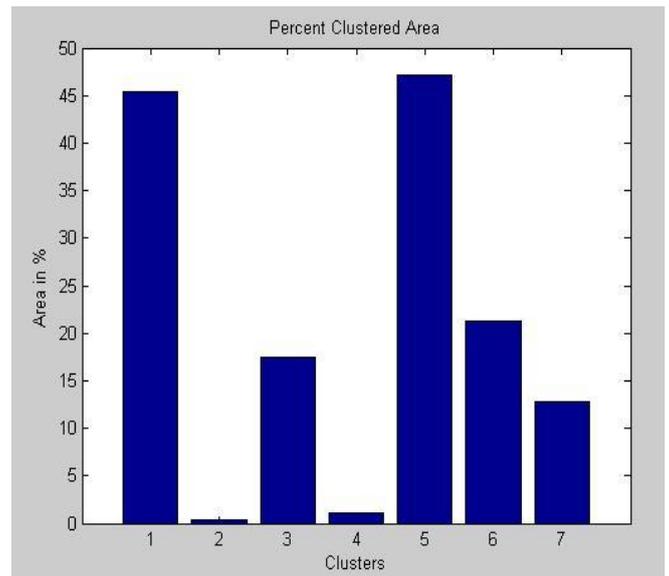

Fig. 6. Percent Area of individual color class clusters (Cluster1 is image background)

The next process of this proposed tool is to visualize all segmented and clustered color class clusters into 3D space feature plot (Isosurface).

Creating an isosurface for each ROI and simultaneously visualizing these in a feature space plot offers two insights. Firstly, the user can examine each training cluster at varying levels of density. This is useful for traditional exploratory data analysis (EDA). Traditionally, visual EDA was used in data mining only as a means of checking that data conformed to assumptions prior to analysis. For the maximum likelihood classification algorithm this means checking the training data for a normal distribution. Thematic classes in satellite imagery often do not conform to assumptions made by classification algorithms. Secondly, the user may explore overlap between training clusters. This is another use of traditional EDA to check underlying assumptions. Many classifiers struggle to deal appropriately with overlapping training data introducing uncertainty in the classification result. It is important to visualize both of these phenomena prior to supervised classification in order to interpret the results such analyses [1].

The study is a simple 6 class problem (excluding cluster1 for 3D isosurface) with training regions as shown in Figure 1. A random sample of 160 pixels was extracted from each training area for classification and a further 160 independently, randomly sampled pixels extracted for accuracy assessment. Visualization of the training regions is performed using all pixels in the regions. Firstly, bands 3, 2 and 1 are selected to be used for classification, and hence visualization. The tool surf() function in MATLAB is configured to display each region (color classes) as an isosurface (sphere) [4].

The result is shown in Figure 7. The surface and sphere functions are used to compute this overlap and display it as a new isosurface. The mean is the only property of the data used by the minimum distance classifier.

Pixels are classified according to the closest mean point in feature space. The visualization functions show that the classified clusters: constructed (red), vegetation (green), water (blue), agriculture (yellow), rocky/barren (gray) and scrub land (brown) area classes in the form of isosurface. The following Table 3 showing the classified values of 6 different classes. And the Figure 7 showing the 3D space feature plot for isosurface of 6 training regions.

Table III. LULC Classified Values of 6 Class Clusters

| Clusters | Area (%) | Pixel Color | | | Freq. (Pixels) |
|---|---|---|---|---|---|
| | | Red | Green | Blue | |
| Cluster 1 | 0.33 | 1 | 0.24706 | 0.26275 | 616 |
| Cluster 2 | 17.45 | 0.60784 | 0.81176 | 0.1451 | 32849 |
| Cluster 3 | 1.13 | 0.42353 | 0.59216 | 0.96471 | 2123 |
| Cluster 4 | 47.14 | 1 | 1 | 0.5098 | 88743 |
| Cluster 5 | 21.23 | 0.94118 | 0.90196 | 0.79608 | 39960 |
| Cluster 6 | 12.72 | 0.80784 | 0.69412 | 0.52941 | 23951 |

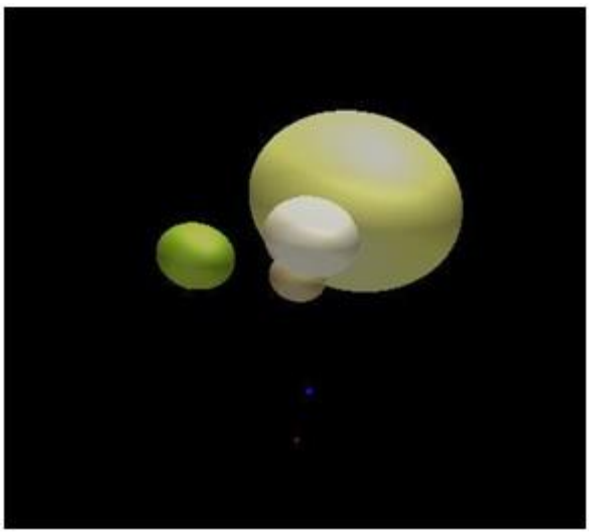

Fig. 7. A 3D Space Feature Plot Showing ISO Surface For Training Regions (Color Class Clusters)

## IV. CONCLUSION

The VDM used to understand the classified objects easily for visual analysis. This study presented the easiest way to visualize 3D classified patterns using visualization with data mining (VDM) techniques and it can provide to analysis of satellite (LULC) imagery. A volume based representation is used for analysis and visualization using different color class clusters. Isosurface and ellipsoids (sphere) where used to construct 3D feature space plots showing the differences between training regions used during classification. This is easiest and fastest method to classify LULC imageries for land assessment, analysis, change detection, planning and development, etc.